\documentclass[twocolumn]{article}
\usepackage{times}
\usepackage{soul}
\usepackage{url}
\usepackage[hidelinks]{hyperref}
\usepackage[utf8]{inputenc}
\usepackage{graphicx}
\usepackage{amsmath}
\usepackage{booktabs}
\usepackage{algorithm}
\usepackage{algorithmic}
\usepackage{amssymb}
\usepackage{stmaryrd}
\usepackage{authblk}
\usepackage[margin=0pt]{subfig}
\title{A New Approach for Explainable Multiple Organ Annotation with Few Data}

\author[1,2]{R\'egis Pierrard}
\author[1]{Jean-Philippe Poli}
\author[2]{C\'eline Hudelot}
\affil[1]{CEA, LIST, 91191 Gif-sur-Yvette cedex, France.}
\affil[2]{Universit\'e Paris-Saclay, CentraleSup\'elec, Math\'ematiques et Informatique pour la Complexit\'e et les Syst\`emes, 91190, Gif-sur-Yvette, France.}
\affil[ ]{\textit{\{regis.pierrard, jean-philippe.poli\}@cea.fr, celine.hudelot@centralesupelec.fr}}
\date{}

\begin{document}

\maketitle

\begin{abstract}
   Despite the recent successes of deep learning, such models are still far from some human abilities like learning from few examples, reasoning and explaining decisions. In this paper, we focus on organ annotation in medical images and we introduce a reasoning framework that is based on learning fuzzy relations on a small dataset for generating explanations. Given a catalogue of relations, it efficiently induces the most relevant relations and combines them for building constraints in order to both solve the organ annotation task and generate explanations. We test our approach on a publicly available dataset of medical images where several organs are already segmented. A demonstration of our model is proposed with an example of explained annotations. It was trained on a small training set containing as few as a couple of examples.
\end{abstract}

%%%%%%%%% BODY TEXT
\section{Introduction}

In the last few years, explaining outputs returned by Artificial Intelligence (AI) algorithms has become more and more important~\cite{RGPD,gunning2017explainable}. This echoes the dominance of deep neural networks, which reach very high performance in several visual recognition tasks but lack of explainability~\cite{marcus,GARNELO201917}. Explaining decisions returned by intelligent systems is not only helpful for understanding their reasoning process, it is also essential for gaining acceptance and becoming trustworthy to humans~\cite{ribeiro2016should}. In human-centered  fields like medical image analysis~\cite{LITJENS201760}, decisions cannot be made relying blindly on a model since the consequences could be disastrous.
%Moreover, it could be an efficient tool for detecting biases in datasets~\cite{torralba}.
%It is especially true in computer vision applications since deep neural networks are able to exceed human performance in several tasks. For instance, the human top-5 classification error rate on the large scale ImageNet dataset equals 5.1\%~\cite{Russakovsky2015}, while best classifiers reach 2.25\%~\cite{hu2017squeeze}. In healthcare applications, explaining and understanding what is going on inside an AI is a paramount concern because they directly deal with human beings~\cite{gunning2017explainable}.

While several definitions of interpretability and explainability exist in the literature~\cite{miller,gilpin,Lipton,DoshiKim2017Interpretability}, there is no consensus among them and these two notions are sometimes used interchangeably. Overall, it emerges that interpretability is the ability to present insight into how a system works in understandable terms, whereas explainability is the ability to describe how a system works in an accurate and logical way. In this paper, we focus on rendering the reasoning process of our model to explain its decisions. To get explanations, a first family of methods consists in learning a local interpretable approximation model around the prediction returned by a black-box model~\cite{shap,ribeiro2016should}. Those approaches can deal with any model, so they are well-suited for deep neural networks. However, although they aim at extracting key characteristics that led to the output, they cannot exactly replicate the reasoning the black-box model performed. The second possibility is to use models that are propitious for generating explanations, such as decision trees, decision rules or by distilling an unexplainable model into an explainable one~\cite{distillation}. Their main advantage is that the reasoning leading to a specific output is easy to track, so it can be used for generating an explanation. However, those models may not be as effective as black-box models, since explainability usually comes at a cost. Indeed, there is a well-known trade-off between accuracy and explainability~\cite{gunning2017explainable}. In this paper, we propose to rely on this second family of approaches by counterbalancing this trade-off with very little need for labelled data whose acquisition is costly. Our approach is based on two conclusions from human image interpretation studies: (1) the importance of contextual and spatial relations in object and scene recognition~\cite{biederman}, and (2) the ability of humans to learn from few examples~\cite{thorpe1996speed,Li}. Several approaches focus on few data learning~\cite{zeroShot,oneShot} but they need side information. We propose to mix statistical and symbolic learning to train a model that learns to manipulate spatial relations from few examples.
%Zero-shot~\cite{zeroShot}, one-shot and few shot learning~\cite{oneShot} aim at doing that relying on previously learned images or available image representations. However, such methods always need side information between seen and unseen classes~\cite{xlsa18}.

Our goal is to build a novel approach that can learn to reason and generate both annotations and explanations from just few examples. In our experiments, the organs to annotate all have properties and they are all linked by spatial relations. Thus, learning these relations and properties should help us to recognize them. Our approach relies on using fuzzy relations that take into account both quantitative and qualitative information, which enables to have a linguistic and thus understandable description of each relation. Learning fuzzy relations has already been proposed in~\cite{donadello} and in~\cite{gonzalez2012efficient} to achieve higher classification performance but not for explaining the reasoning as we propose. Given an unknown example, the system looks for the set of objects that best satisfies the relations between the objects of interest. We model this as a constraint satisfaction problem. In Section~\ref{sec:proposedApproach}, we describe the whole pipeline that consists in three main steps: assessing relations, extracting the most relevant ones and generating constraints for solving a constraint satisfaction problem and producing explanations.
%First, given a vocabulary, relations are assessed according to a strategy that enables the system to first evaluate the most generic ones. Then, the most relevant relations are induced. Those are the relations that will be used for annotating and explaining. Finally, constraints are generated to solve a fuzzy constraint satisfaction problem and to provide explanations to the solution.
%
%Annotating organs in medical images is helpful for saving medical staff time, but also for being used as a priori information in classification and segmentation tasks. This has hardly been tackled in the literature~\cite{xue2017automatic}. Organ classification~\cite{roth2015anatomy} and organ localization~\cite{criminisi,hooChangShin} are more common topics. They rely on deep neural networks or regression forests, which are not suited for generating explanations. Our work aims at solving this problem while providing an explanation for it performing visual reasoning.
%
In Section~\ref{sec:experiments}, a demonstration of this approach is shown on a task of multiple organ recognition on medical images. This task is a good example of spatial reasoning since the spatial arrangement of the organs plays an important role in their recognition. In addition, working on medical images presents several challenges, including a need for explainability and the fact that datasets are usually small. We tested and compared our model to the state of the art and showed that our approach is able to achieve high accuracy and generate explanations in spite of a low number of training data.

\section{Background}

The approach we present in the next section relies on learning relevant fuzzy relations between objects for defining a constraint satisfaction problem. All the notions that are involved are reminded in this section.

\subsection{Fuzzy Logic}

Fuzzy logic and fuzzy set theory~\cite{ZADEH1965338} can be seen as an extension of Boolean logic that enables to manage imprecision. %While a value is either true or false in Boolean logic, it can take any value from 0 (false) to 1 (true) in fuzzy logic.% This range of degrees is useful for dealing with vagueness. %For example, natural language expressions can be represented and so it may help to make a system interpretable.
In a universe $A$, a fuzzy set $F$ is characterized by a mapping such as $\mu_F : A \rightarrow [0,1]$. This mapping specifies in what extent each $a \in A$ belongs to $F$ and it is called \emph{the membership function of} $F$. If $F$ is a non-fuzzy set, $\mu_F(a)$ is either 0, i.e. $a$ is not a member of $F$, or 1, i.e. $a$ is a member of $F$. This range of degrees is useful for dealing with vagueness.
%The set of all fuzzy sets in a universe $X$ is written $F^X$.

% The concept of linguistic variable~\cite{zadeh1975concept} can then be defined as a triplet $(V, X, T_V)$ such as:
% \begin{itemize}
% 	\item $V$ is the name of the variable (e.g. ``distance'');
%     \item $X$ is the domain on which $V$ is defined (e.g. $\mathbb{R}$ or $[0,100]$);
%     \item $T_V = \{T_1, T_2,...\}$ is a finite collection of fuzzy sets. Each of these fuzzy sets is associated to a linguistic term which qualifies $V$ (e.g. ``short'' or ``long'').
% \end{itemize}

% An elementary fuzzy proposition ``$V$ is $A$'' is defined from a linguistic variable $(V, X_V, T_V)$ with $A \in T_V$. For example, the fuzzy proposition ``distance is short'' is assessed using the membership function $\mu_{short}$. For a specific distance $d$, the truth value of this proposition is returned by $\mu_{short}(d)$. This truth value is between 0 and 1 and represents how short the distance $d$ is. For instance, it is interesting to note that $\mu_{short} = 0$ only means that $d$ is not short at all. But it does not necessarily mean that $d$ is long since there can be several intermediary linguistic terms in $T_V$ between ``short'' and ``long''. The whole framework enables to have a rich vocabulary which contributes to make the system interpretable.

The fuzzy logic framework is also convenient for expressing relations between two sets. Given two universes $A$ and $B$, a binary fuzzy relation $\mathcal{R}$ is characterized by a mapping defined as $\mathcal{\mu_R} : A \times B \rightarrow \left[0,1\right]$. It assigns a degree of relationship to any $(a,b) \in A \times B$. $n$-ary fuzzy relations are defined identically. Another advantage is that fuzzy logic allows using words instead of mathematical symbols. %When associated to a linguistic description, which is common in fuzzy logic, a fuzzy set or a fuzzy relation can characterize a natural language expression.

% In the following, for the sake of comprehension, the word \emph{relation} refers either to a fuzzy relation $R$ or to the degree of relationship $\mu_R(x,y)$ between $x$ and $y$. For instance, for a binary fuzzy relation $R_{\textrm{to the left of}}$ and two objects $a$ and $b$, we call relation the result $\mu_{R_{\textrm{to the left of}}}(a,b)$, which represents the spatial relation ``$a$ to the left of $b$''.

% \begin{figure}[tb]
% 	\centering
%     \includegraphics[width=0.40\textwidth]{figures/dog}
%     \caption{Example of dog classification~\protect\cite{SimonyanVZ13} which illustrates the difference between interpretation and explanation. On the left, the attention map is an interpretation because it is easy to understand relatively to the input but it does not describe the logic that led to the classification. On the right, it is an example of explanation since the causes are clearly stated.}
%     \label{fig:dog}
% \end{figure}

\subsection{Fuzzy Constraint Satisfaction Problem}
\label{subsec:back}

A constraint satisfaction problem (CSP) %~\cite{MACKWORTH197799,montanari,waltz1972generating}
consists in assigning some values to a set of variables that must respect a set of constraints, such as scheduling problems~\cite{MINTON1992161} for instance. \cite{Dubois1996} presents an extension of CSPs to the fuzzy logic framework to deal with imprecise parameters and flexible constraints. This is called a fuzzy constraint satisfaction problem (FCSP). A FCSP is defined by a set of variables $X$, a set of domains $D$ and a set of flexible constraints $C$. It is an appealing framework in the context of explainable annotation since it enables to both solve the annotation task (getting each variable assignment) and generate explanations using the constraints.

% One instantiation that satisfies the FCSP is evaluated by its degree of consistency. As finding consistent solutions is too complex, it is usually more convenient to use only local consistency, which leads to a simpler problem. One of the most used consistencies is arc-consistency (or 2-consistency). A famous filtering algorithm based on arc-consistency is AC3~\cite{MACKWORTH197799} and its extension to FCSPs is called FAC-3~\cite{Dubois1996}. \cite{VANEGAS20161} presents an updated version of this algorithm that can deal with groups of objects and $n$-ary flexible constraints. The principle of the FAC-3 algorithm is to check all the constraints in the FCSP to determine which domains are consistent for every variable.

To solve a FCSP, the FAC-3 algorithm~\cite{Dubois1996,VANEGAS20161} is usually applied to prune the search space. Then, a backtracking algorithm explores every possible solution. Finally, we get the best solution by picking the one that is the most consistent with the set of constraints $C$.

% \subsection{Interpretability and Explainability}

% Since explainability is one of the purpose of the approach we present in this paper, it is especially important to define it. Explainability issues were already tackled in the 70's by expert systems~\cite{feigenbaum1970generality,SHORTLIFFE1975351}. Now they are highlighted again~\cite{RGPD,gunning2017explainable} and explainability has become one of the main area of development in the machine learning community~\cite{2018arXiv180701308K}. However, there is no reference definition and the difference between interpretability and explainability can be vague. Indeed, those notions are often used to define each other while they are not exactly the same~\cite{DoshiKim2017Interpretability}.

% Explainability implies understanding the reasoning that led to a decision whereas interpretability consists in giving clues how a system works in a way that is understandable to humans. Figure~\ref{fig:dog} illustrates this: the attention map on the left is a visual clue that can help understand what was going on inside the CNN that performed the classification. It is more intuitive than the explanation on the right, but it does not tell much about the reasoning that was performed. The approach we present here aims at returning an explanation an not an interpretation.

\section{Proposed Approach}
\label{sec:proposedApproach}

\begin{figure}[tb]
    \centering
    \includegraphics[width=0.5\textwidth]{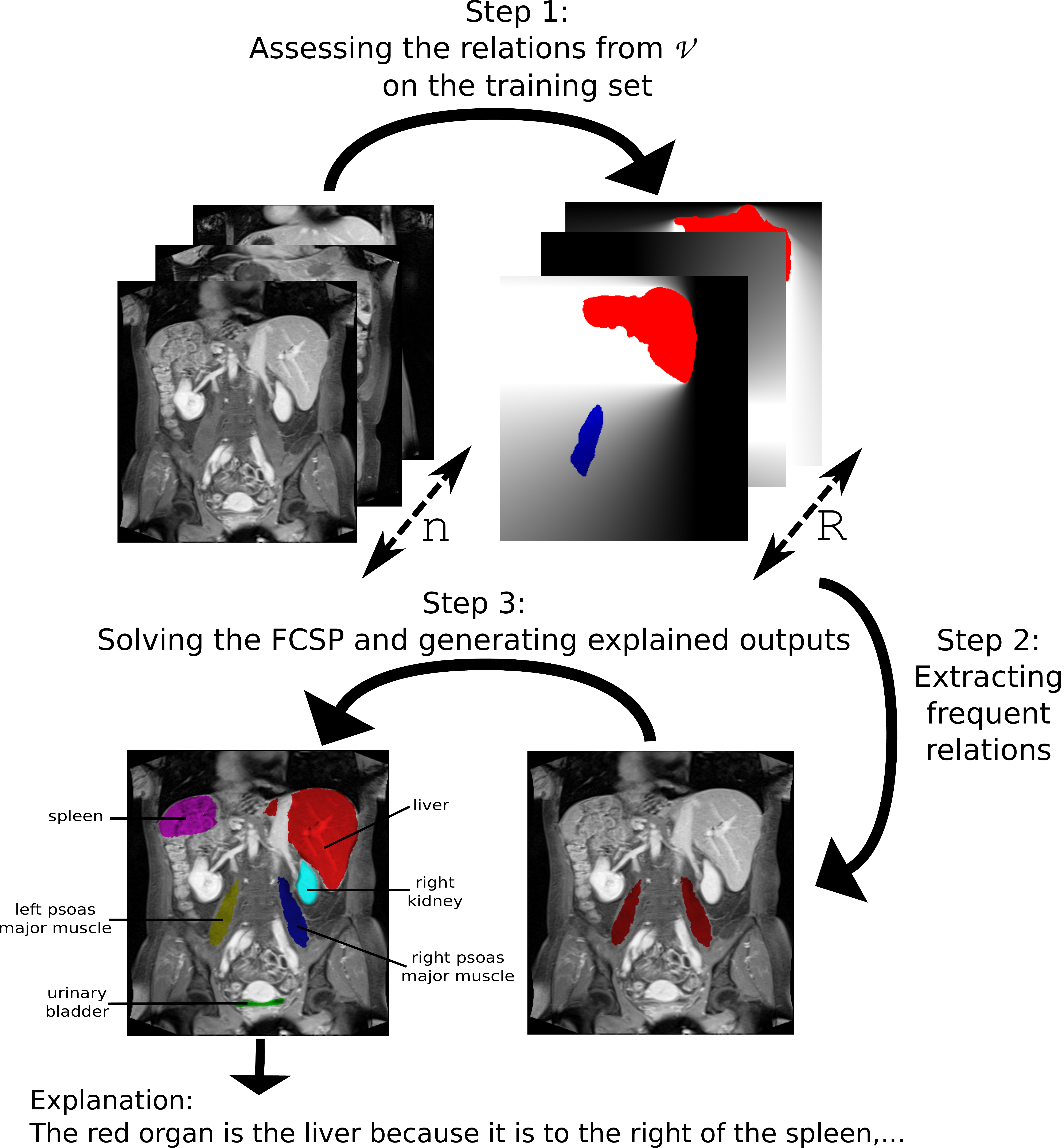}
    \caption{Illustration of our explainable multiple organ annotation system. In step~1, the $R$ fuzzy relations from the vocabulary $\mathcal{V}$ are evaluated on a training set of $n$ images. In step~2, the most frequent of them are extracted to set constraints. In step~3, for each test image, a FCSP is defined and solved to label the different regions. An explanation is provided for each labeling based on the constraints that are used.}
    \label{fig:overview}
\end{figure}

In this section, we describe our new approach that aims at annotating regions of interest in images and at providing an explanation for each annotation. It consists of three steps: the assessment of fuzzy relations from a given vocabulary between the organs we are looking for (Sec.~\ref{subsec:step1}), the learning of the most relevant relations between the organs (Sec.~\ref{subsec:mining}) and the solving of a FCSP providing explanations for finding the regions that are the most consistent with the relevant relations and explaining the reasoning behind it (Sec.~\ref{subsec:expl}). An overview of the whole approach is illustrated in Figure~\ref{fig:overview}.

\subsection{Step 1: Assessing Relations}
\label{subsec:step1}

This step aims at evaluating several relations between the regions of interest (the organs) so that we can later (in the following step) find the most relevant of them.

Let us consider a training set $\mathcal{T_{\textrm{train}}}$ that contains $n$ images $\{\textbf{i}_1, \dots, \textbf{i}_n\}$ and a set of labels $\mathcal{Y}$ that contains $N$ labels $\{y_1, \dots, y_N\}$ such as each image $\textbf{i} \in \mathcal{T_{\textrm{train}}}$ is divided into $K$ regions of interest $\{o_{\textbf{i}, 1}, \dots, o_{\textbf{i}, K}\}$ that are mapped to labels by the following function:
\begin{equation}
\begin{alignedat}{3}
    f : \{o_{\textbf{i}, 1}, &\dots, o_{\textbf{i}, K}\} & \rightarrow & \mathcal{Y}\\
    & o_i & \mapsto & y_j
\end{alignedat}
\end{equation}

Let us consider a set $\mathcal{V} = \{\mathcal{R}_1, \dots, \mathcal{R}_R\}$ of relations. We call this set a vocabulary. It is set by an expert in the target task and it is composed of would-be relevant relations. For example, one relation can be a directional relation like \emph{to the left of} or a distance relation like \emph{close to}. The richer the vocabulary, the more expressive the system which should help to produce better annotations and explanations. Relations in $\mathcal{V}$ are automatically evaluated on the regions of interest of each image in $\mathcal{T_{\textrm{train}}}$. The way they are computed depends on the definition of the relation, as shown in Sec.~\ref{subsubssec:relations}.

For any relation $\mathcal{R} \in \mathcal{V}$, let $\alpha(\mathcal{R})$ denote its arity. $\mathcal{R}$ is evaluated for each possible $\alpha(\mathcal{R})$-tuple of regions of interest. It is important to distinguish $\mathcal{R}$ from its evaluations on the different regions. The number of evaluations to perform is:
\begin{equation}
    \sum\limits_{p=1}^{n}\sum\limits_{j=1}^{R}\frac{K!}{(K-\alpha(\mathcal{R}_j))!}
\end{equation}

At the end of this step, we have a set of evaluated relations between organs $\{\mathcal{R}(f(o_{\textbf{i}, v}), f(o_{\textbf{i}, w})) \mid \mathcal{R} \in \mathcal{V}, \textbf{i} \in \mathcal{T_{\textrm{train}}}, o_{\textbf{i}, v}, o_{\textbf{i}, w} \in \textbf{i}\}$ that can be seen as features.

\subsection{Step 2: Learning Relevant Fuzzy Relations}
\label{subsec:mining}

In this step, the objective is to extract among the previously assessed relations the most relevant of them.
For a label $y \in \mathcal{Y}$, our postulate is that the relevant relations involving the regions labelled as $y$ are the most frequent ones since they should be verified by most, if not all, examples of these regions. Thus, learning the relevant relations is performed by mining the most frequent ones. It is done in a one-vs-all way since the relevant relations for one class of organs are not the same as for a different class. As each example from one class should be correlated to each other, we use a fuzzy mining algorithm that takes advantage of that~\cite{close}.

Let $E(\mathcal{V})$ be the set of all the evaluations of relations from $\mathcal{V}$ on the labeled regions of interest. A subset of relations $J$ is a set belonging to $2^{E(\mathcal{V})}$. The mining algorithm we use is based on a fuzzy closure operator $h : 2^{E(\mathcal{V})} \rightarrow 2^{E(\mathcal{V})}$ that enables to find all the closed sets of relations~\cite{close}. All the frequent closed sets of relations are computed and the frequent sets of relations can be derived from them. A set of relations is said to be frequent when its frequency in the dataset is larger than a given threshold. Since this step is performed in a one-vs-all way, each class has its own threshold whose value is an hyperparameter determined during a validation phase.
%The fuzzy algorithm~\cite{close} that is used in step~2 requires a threshold to be set for assessing the frequency of a subset of relations. As the training is performed in a one-vs-rest way, this threshold needs to be reset for each class. That means that the number of hyperparameters of the model is equal to the number of classes.
The value of this threshold has a direct impact on the number of frequent subsets of relations that are extracted. If it is too high, it is likely that no or few subsets of relations are seen as frequent, which may be not enough for discriminating classes. This would be a case of underfitting. On the other hand, if the threshold is too low, some irrelevant features will be kept. That would lead to overfitting. At the end of this step, for each label $y \in \mathcal{Y}$, we have a set of frequent subsets of evaluated relations $F_{y}$ such as $F_{y} \subseteq 2^{2^{E(\mathcal{V})}}$.

\subsection{Step 3: Solving the FCSP and Generating Explanations}
\label{subsec:expl}

Given a test example $\textbf{i}$, we can obtain a set of potential regions of interest by segmentation. The goal of this step is to find the labels of the regions that best satisfy the relations between organs that were learnt in the previous step. This can be modelled as a FCSP. Also, since these relations are associated to a linguistic description, we can generate an explanation for each annotation.

For each label $y \in \mathcal{Y}$, we got at the end of the previous step a set $F_y$. Let us define $F_{y}^{\max}$ such as :
\begin{equation}
    F_{y}^{\max} = \{z \in F_{y} \mid \textrm{Card}(z) = \max\limits_{v \in F_{y}}\big(\textrm{Card}(v)\big)\}
\end{equation}
This set corresponds to the set of the frequent subsets of relations of maximal size.
Each evaluated relation $\mathcal{R}\big(f(o_{\textbf{i}, v}), f(o_{\textbf{i}, w})\big)$ in the subsets of relations in $\bigcup\limits_{y \in \mathcal{Y}}F_{y}^{\max}$ is directly translated into a constraint $c_{\mathcal{R}}\big(f(o_{\textbf{i}, v}), f(o_{\textbf{i}, w})\big)$. We can now build a model that is defined by the constraints that have been learned and its frequency thresholds. No iterative optimization process is needed, which makes it well suited to small training sets.

The test example $\textbf{i}$ is divided into $K$ regions of interest $\{o_{\textbf{i}, 1}, \dots, o_{\textbf{i}, K}\}$ that we want to annotate. The FCSP we get is the following :
\begin{equation}
    X = \{o_{\textbf{i}, 1}, \dots, o_{\textbf{i}, K}\}
\end{equation}
\begin{equation}
    D = \{D_j \mid D_j = \mathcal{Y}, 1 \leq j \leq K\}
\end{equation}
\begin{multline}
    C = \{c_{\mathcal{R}}(f(o_{\textbf{i}, v}), f(o_{\textbf{i}, w})) \mid \mathcal{R}(f(o_{\textbf{i}, v}), f(o_{\textbf{i}, w})) \in U \\ \textrm{ such as } U \subseteq \bigcup\limits_{y \in \mathcal{Y}}F_y^{\max}\}
\end{multline}

Then, each constraint in $C$ is evaluated, the FCSP is solved and the first part of the output, the labels, are returned. We obtain a new mapping $f_{\textbf{x}}$ such as :
\begin{equation}
\begin{alignedat}{3}
    f_{\textbf{i}} : \{o_{\textbf{i}, 1}, &\dots, o_{\textbf{i}, K}\} & \rightarrow & \mathcal{Y}\\
    & o_i & \mapsto & y_j
\end{alignedat}
\end{equation}
Then, for each variable $o_{\textbf{i}, j} \in X$, an explanation is generated using the constraints in $C$. This is possible because the relations (and so the constraints) that we use are associated to a linguistic description. 
%One of the main characteristics of our approach is that it is able to provide an explanation for the results it returns. Generating these explanations relies on the constraints that have been learned on the training set.
%A constraint can be represented as a couple $(E, \mathcal{R})$ with $\mathcal{R}$ a $\alpha(\mathcal{R})$-ary relation and $E$ a set of $\alpha(\mathcal{R})$ regions that are involved in $\mathcal{R}$.
For instance, the constraint $c_{\mathcal{R}_{\textrm{to the left of}}}(A, B)$ (represented as a tuple $(A, B, \mathcal{R}_{\textrm{to the left of}})$) leads to: ``$A$ is to the left of $B$''. Thus, using the constraints generated from $F_{y}^{\max}$ enables to express an explanation in the form of ``\emph{output} BECAUSE \emph{cause}$_1$,...,\emph{cause}$_n$''. For a given label $y$, all the constraints related to $y$ are extracted. The least satisfied constraint gives us a certainty factor to moderate the explanation~\cite{Budescu2012}, e.g. ''\emph{This organ is likely to be annotated as the liver...}``. The constraints and the certainty factor are then sent to a surface realiser like simpleNLG~\cite{simplenlg} to aggregate them into a syntactically correct sentence.
%In parallel, we work on an improvement of the formulation of explanations based on natural language generation techniques but we are not focusing on it here.

\section{Case study}
\label{sec:experiments}

In this section, we detail the experiments we have performed on a dataset of medical images. The task is to perform explained multiple organ annotation by learning a model from few data. While multiple organ detection has been a regularly tackled topic in the literature~\cite{hooChangShin,criminisi,pauly,lee}, multiple organ annotation has only been tackled in \cite{xue2017automatic}. The principle of this method is to find images in the dataset that share visual characteristics with the image under study, and then to label it based on the labels from visually similar images. However, it cannot provide any explanation. In \cite{lee}, abdominal organ detection is performed using fuzzy spatial rules, but these rules are not suited to other datasets and they have to be set by an expert before learning. Organ classification has been addressed in~\cite{roth2015anatomy} using data augmentation to dodge the problem of having a small training set.

\subsection{Dataset}
\begin{figure}[tb]
	\centering
	\subfloat[CTwb]{\includegraphics[height=2.8cm]{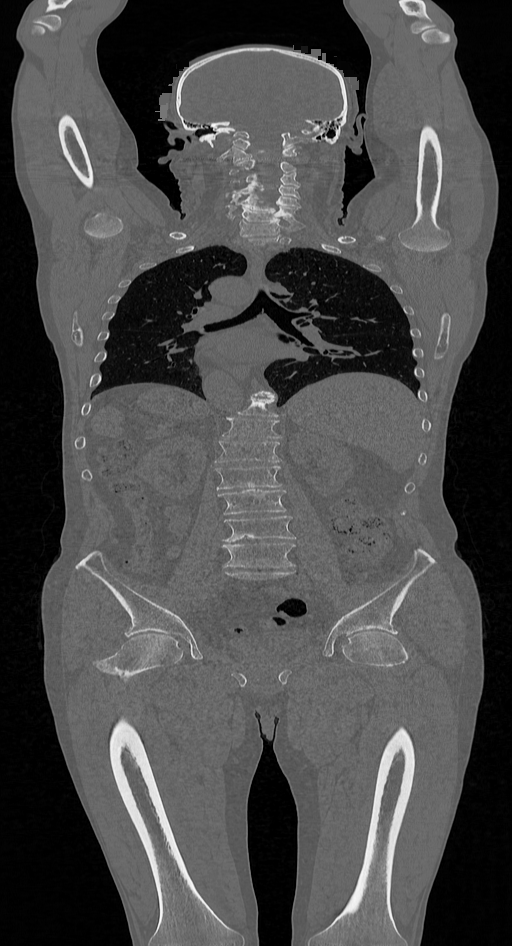}}
    \hfill
    \subfloat[CTce]{\includegraphics[height=2.8cm]{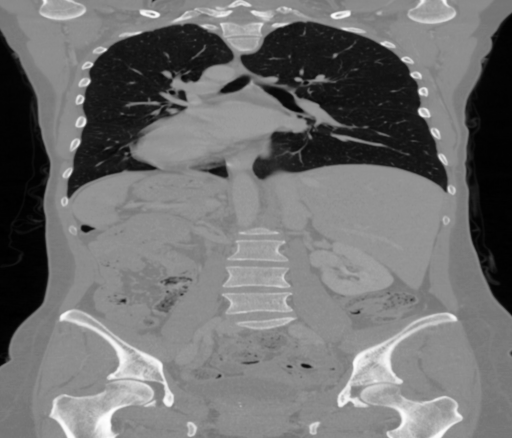}}
    \hfill
    \subfloat[MRwb]{\includegraphics[height=2.8cm]{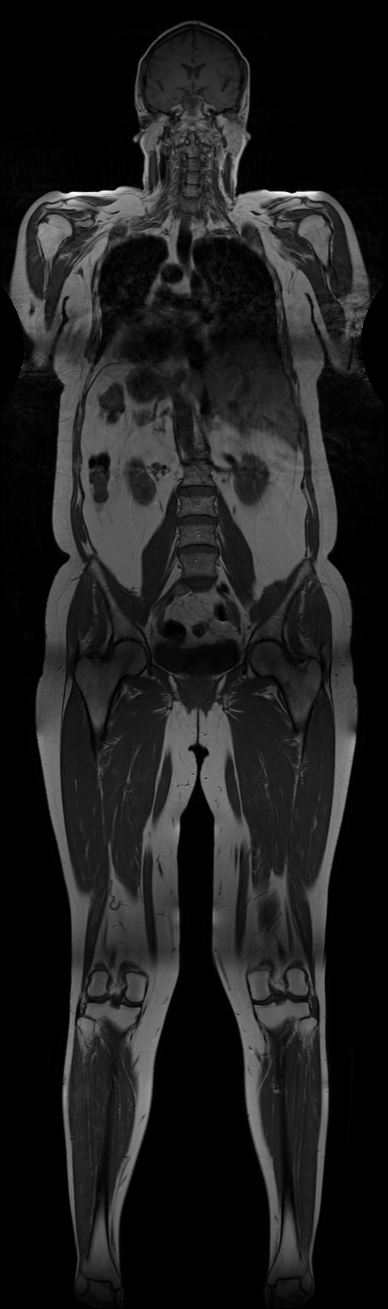}}
    \hfill
    \subfloat[MRce]{\includegraphics[height=2.8cm]{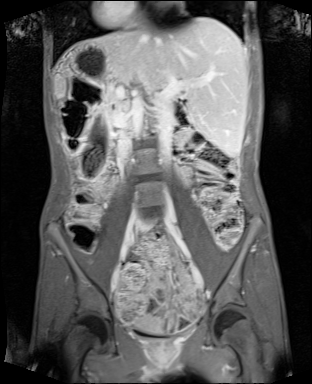}}
    \caption{Examples of the four types of scans in the dataset.}
    \label{fig:visceral}
\end{figure}

It is important to note that the field of XAI is currently lacking a dataset that mainly focuses on explanations. This is why we carried out our experiments on a segmentation dataset that we used for assessing the accuracy of our model and the reliability of the explanations it produces. This dataset is named \emph{Anatomy3} and has been presented in~\cite{jimenez2016cloud}. It contains 391 CT and MR images and their corresponding segmented organs. Images can be scans of the whole body (referred as CTwb and MRwb) or enhanced images of the abdomen (referred as CTce and MRce). Those are all 3D images that are actually the superposition of 2D slices. As we work on 2D images, we consider only slices in the following. We selected the slices to build a 2D image dataset. Figure~\ref{fig:visceral} displays one example for each type of scan.

The set $\mathcal{Y}$ of organs (labels) we study is composed of the \emph{liver}, the \emph{spleen}, the \emph{urinary bladder}, the \emph{left} and \emph{right kidneys}, the \emph{left} and \emph{right lungs} and the \emph{left} and \emph{right psoas major muscles}. We kept all the images that contain these 9 organs (and their corresponding segments), for a total of 35 examples and 315 segments in our dataset.

% \begin{figure}[p]
% 	\centering
%     \includegraphics[width=\textwidth]{Figures/sample.pdf}
%     \caption{Only the colored organs are considered in our dataset. The 4 red organs are the ones we would like to label, while the blue organs are used to make the number of domains larger.}
%     \label{fig:organs}
% \end{figure}

\subsection{Experimental Settings}
\subsubsection{Model Training}

The model we build with our approach consists in the frequent subsets of relations that are extracted. There are as many hyperparameters as labels and they correspond to the thresholds used for assessing the frequency of a subset of relations. Model selection is necessary to get optimized thresholds, which is why we used \emph{nested cross-validation}~\cite{cawley2010over}: (1) an outer cross-validation is performed in which we get a training set and a test set for each iteration, (2) an inner cross-validation is performed on the training set of the outer cross-validation to get an inner training set and a validation set for tuning hyperparameters. This enables to get an unbiased error prediction.
%The outer cross-validation is actually a 3-fold cross-validation and the inner one is a 4-fold cross-validation.

In the inner cross-validation, hyperparameter tuning is performed using bayesian optimization over 20 iterations with a Gaussian process prior. The acquisition function is the expected improvement.

\subsubsection{Relations}
\label{subsubssec:relations}

\begin{figure}[tb]
	\centering
    \subfloat[Input]{\label{subfig:input}\includegraphics[width=0.15\textwidth]{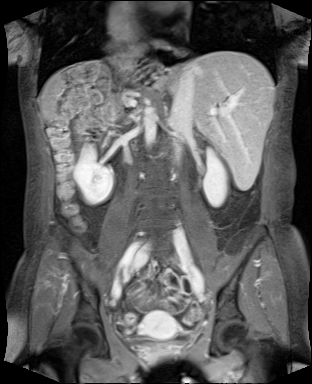}}
    \hfill
    \subfloat[Segmented organ]{\label{subfig:disk}\includegraphics[width=0.15\textwidth]{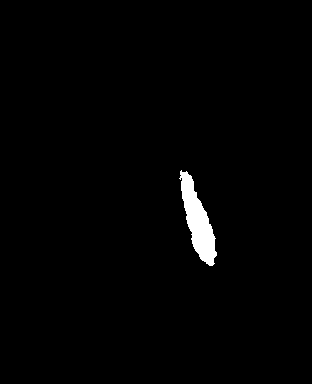}}
    \hfill
    %\subfloat[\emph{To the left of} disk]{\label{subfig:rightDisk}\includegraphics[width=0.12\textwidth]{gauche}}
    %\hfill
    \subfloat[Relation]{\label{subfig:ellipseRightDisk}\includegraphics[width=0.15\textwidth]{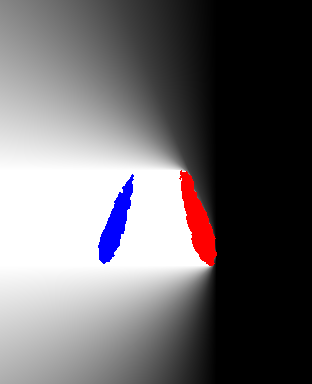}}
    \caption{(Best viewed in color) Example of how an input is used to compute a specific relation. Here, the goal is to compute the relation \emph{blue organ to the left of the red organ}. Given an input (\ref{subfig:input}), a segmented organ is considered (\ref{subfig:disk}) as the reference object. This organ is used to compute a fuzzy landscape (\ref{subfig:ellipseRightDisk}) that represents the degree to which each pixel verifies the relation \emph{to the left of the red organ}. Finally, the relation is assessed by evaluating the degree of intersection between this fuzzy landscape and the blue organ.}
    \label{fig:fpm}
\end{figure}

Many fuzzy spatial relations have been studied in the literature~\cite{bloch2005fuzzy}. In our experiments, we use directional, distance and symmetry relations. Directional and distance relations~\cite{bloch1999fuzzy,bloch1999fuzzy2} are computed as a fuzzy landscape and assessed using a fuzzy pattern matching approach~\cite{cayrol1982fuzzy}. As shown in Figure~\ref{fig:fpm}, the fuzzy landscape is generated by computing the fuzzy morphological dilation of a reference object by a structuring element whose shape determines the kind of relation. Let $S$ be the space of the images. Let $A$ be a reference object in $S$ and $\mu_{A,\mathcal{R}}$ the membership function associated to the fuzzy landscape representing the relation $\mathcal{R}$ whose reference object is $A$. Let $\mu_B$ be the membership function corresponding to an object $B$ in $S$. The relation $\mathcal{R}$ between $A$ and $B$ is the result of the fuzzy degree of intersection $\mu_{int}$ between $\mu_{A,\mathcal{R}}$ and $\mu_B$ such as~\cite{bloch2005fuzzy}
\begin{equation}
    \mu_{int}(\mu_{A,\mathcal{R}}, \mu_B) = \frac{\sum\limits_{x \in S}\min\Big(\mu_{A,\mathcal{R}}(x), \mu_B(x)\Big)}{\min\Big(\sum\limits_{x \in S}\mu_{A,\mathcal{R}}(x), \sum\limits_{x \in S}\mu_B(x)\Big)}
\end{equation}
For instance, in Figure~\ref{fig:fpm}, the relation $\mathcal{R}$ is \emph{to the left of}, the reference object $A$ is the red organ and the object $B$ is the blue organ.

To get a finite catalogue of relations, we constrained the parameters of these relations to express only relations such as \emph{above} or \emph{close to}.
%We are working on defining a method to set these parameters automatically in the first step of the workflow.

The symmetry relation~\cite{colliot} we use consists in finding the line that maximizes a symmetry measure between two organs. Since this measure is not differentiable, a direct search method is used to solve this optimization problem, such as the downhill simplex method.

We also use one property that can be seen as a unary relation since it characterizes just one organ. It evaluates how stretched an organ is. Given a segmented organ, a PCA is performed to get its two principal axes. Then, the organ is projected on both axis and the ratio of these projections is used to compute the degree corresponding to this property. However, this does not manage concave shapes well.

%\subsubsection{Expressivity}

Our vocabulary of relations $\mathcal{V}$ contains: \emph{to the left of}, \emph{to the right of}, \emph{below}, \emph{above}, \emph{close to}, \emph{symmetrical to} and \emph{stretched}. That makes 6 binary and one unary relations. As we consider 9 organs, the number of relations to evaluate for one image is equal to 441, which contributes to make our model expressive. There is however a trade-off between the expressivity of the system and the computation time needed for assessing all these relations. %Further work will focus on a strategy to avoid computing all the relations but only a preselection of them.

\subsection{Problem initialization}

As stated in in Sec.~\ref{sec:proposedApproach}, the whole process consists in three main steps. The inputs we deal with are segments provided in the datasets. They are not fuzzy, but the process is exactly the same whether we deal with fuzzy or crisp objects. 

The intermediary goal is to generate constraints for defining a FCSP. Once solved, the FCSP returns the labels and constraints are used for generating explanations.

The variables are the segments provided in the dataset. Each of them corresponds to an organ. We have the following FCSP:
\begin{multline}
	X = \{o_{\textrm{liver}}, o_{\textrm{spleen}}, o_{\textrm{bladder}}, o_{\textrm{r\_kidney}}, o_{\textrm{l\_kidney}}, o_{\textrm{r\_lung}}, \\ o_{\textrm{l\_lung}}, o_{\textrm{r\_psoas}}, o_{\textrm{l\_psoas}}\}\nonumber
\end{multline}
\begin{multline}
	D = \{D_{\textrm{liver}}, D_{\textrm{spleen}}, D_{\textrm{bladder}}, D_{\textrm{r\_kidney}}, D_{\textrm{l\_kidney}}, D_{\textrm{r\_lung}}, \\ D_{\textrm{l\_lung}}, D_{\textrm{r\_psoas}}, D_{\textrm{l\_psoas}}\}\nonumber
\end{multline}
where $D_i$ is equal to $\mathcal{Y}$. For each organ $y$, the flexible constraints are generated from the set of the frequent subsets of relations of maximal size $F_{y}^{\max}$ to build a set of constraints $C$. Furthermore, since every organ is unique, there cannot be identical annotations in this problem. That means $C$ has to be extended with constraints representing that two variables cannot be the same, which is the \emph{AllDifferent} global constraint.

The definition of the FCSP is thus made automatically. Then, once the FCSP is defined, for a given example, it can be solved as described in Sec.~\ref{subsec:back}.

\subsection{Results}
%\subsubsection{An Example}

\begin{figure}[tb]
    \centering
    \includegraphics[width=0.36\textwidth]{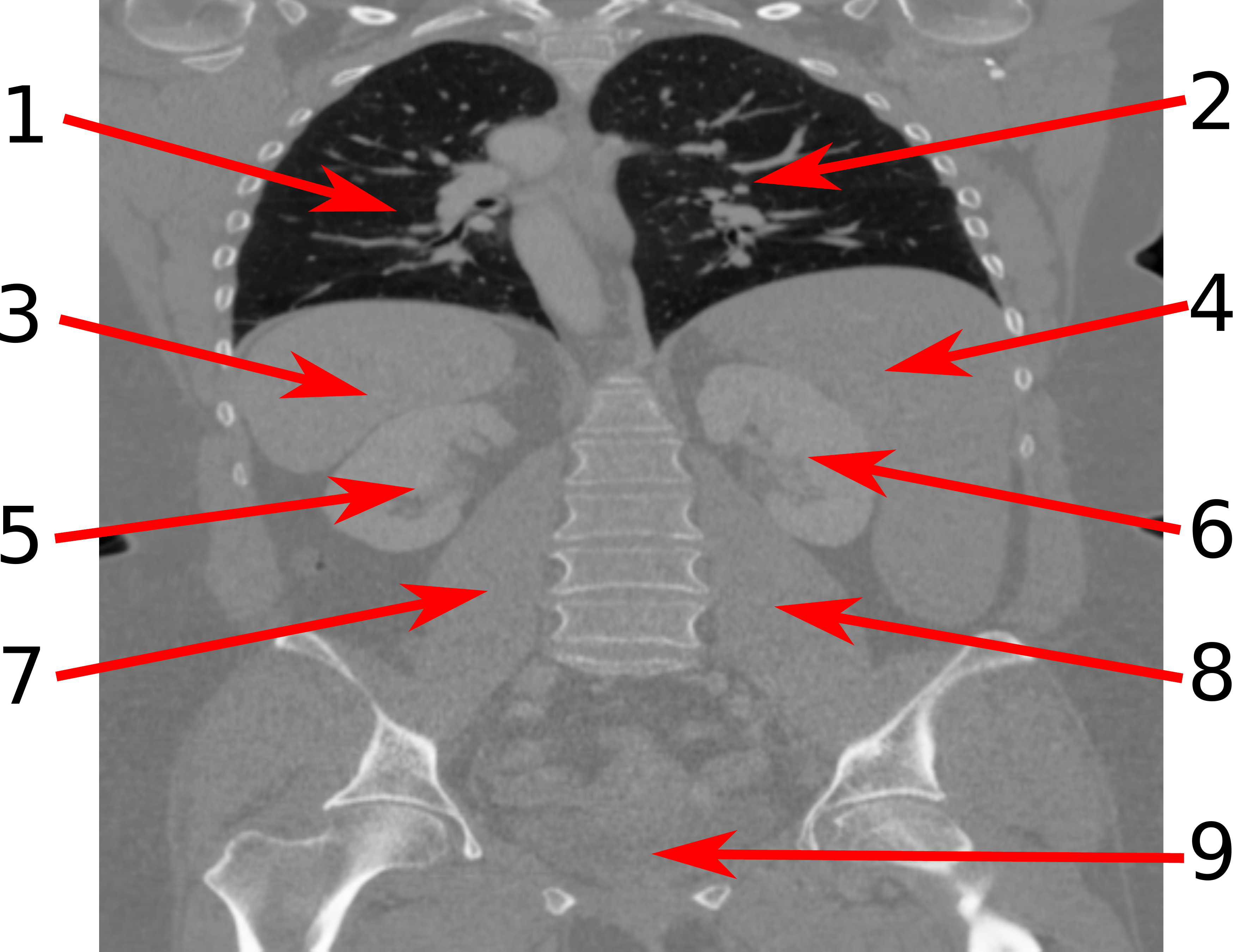}\\
    \footnotesize
    Organ 1 is very likely to be annotated as the left lung \textbf{because} it is \emph{to the left of} the right lung (organ 2), it is \emph{symmetrical} to the right lung and it is \emph{above} the spleen (3).\\
    Organ 9 is likely to be annotated as the bladder \textbf{because} it is \emph{stretched}, it is \emph{below} the right kidney (6) and \emph{below} the left kidney (5).\\
    Organ 4 is very likely to be annotated as the liver \textbf{because}...
    \caption{Example of explained annotations.}
    \label{fig:ex}
\end{figure}

Fig.~\ref{fig:ex} shows an example of output for an input image with 9 organs to annotate and thus 9 explanations to provide.

\begin{table}[tb]
\centering
\resizebox{0.4\textwidth}{!}{%
\begin{tabular}{@{}lc@{}}
\toprule
Organ                                     & Value of the corresponding threshold \\ \midrule
Liver                                   & 0.96          \\
Spleen                        & 0.86          \\
Bladder            & 0.80          \\
Right kidney & 0.92          \\
Left kidney & 0.89          \\
Right lung & 0.98          \\
Left lung & 0.97          \\
Right psoas muscle & 0.92          \\
Left psoas muscle & 0.88          \\ \bottomrule
\end{tabular}%
}
\caption{Values of the thresholds that are the hyperparameters of our model. Each threshold is associated to one organ.}
\label{tab:thresh}
\end{table}

We evaluate our model using the accuracy, which is the ratio for all organs of the number of correct annotations over the total number of annotations. We got an accuracy of $100\%$ for a model containing only directional relations. The outer cross-validation is actually a 3-fold cross-validation (23/24 training examples for 12/11 test examples in each iteration) and the inner one is a 4-fold cross-validation. As there are 9 organs to annotate, there are 9 hyperparameters that need to be set for extracting frequent relations (Table~\ref{tab:thresh}). Constraints could be added to the hyperparameter optimization process to make explanations longer or shorter.
We observe the explanations rightfully rely on the relations that have been extracted and later turned into constraints. For example in Fig.~\ref{fig:ex}, the set of constraints associated to the right kidney is:\\
$C_{r\_kidney} = \{(x_{r\_kidney}, x_{l\_kidney}, \mathcal{R}_{\textrm{symmetrical to}}),$\\ $(x_{r\_lung}, x_{r\_kidney}, \mathcal{R}_{\textrm{above}}), (x_{r\_kidney}, x_{liver}, \mathcal{R}_{\textrm{to the left of}}),$\\ $(x_{bladder}, x_{r\_kidney}, \mathcal{R}_{\textrm{below}}), (x_{r\_kidney}, x_{l\_kidney}, \\ \mathcal{R}_{\textrm{to the right of}}),$ $(x_{l\_kidney}, x_{r\_kidney}, \mathcal{R}_{\textrm{to the left of}})\}$.
Some of these constraints may seem redundant, like the last two constraints in $C_{r\_kidney}$. That can happen because fuzzy morphological dilations depend on the shape of the reference object. As two different organs are never exactly the same, there are slight differences between those two constraints. Each organ is linked to such a set of constraints. The final set of constraints $C$ is the union of all these sets.

Assessing the quality of the explanations is tricky. What makes a good explanation ultimately depends on the knowledge and expectation of the end-user. Criteria like the coherence, the simplicity and the relevancy of the explanation are good indicators~\cite{miller,baaj}, but they may not be easy to assess. Three evaluation methods are proposed in~\cite{DoshiKim2017Interpretability}: asking an expert, asking simple questions to a group of non-expert people or using a proxy model that has been proved to be explainable to assess the model under study.

%Although the model trained with directional relations does not need other types of relations to achieve perfect accuracy, more relations are useful to build better explanations. As shown in Figure~\ref{fig:ex}, the symmetrical relation is used in the explanation of the annotation of the left lung. This is not necessary to get a $100\%$ accuracy, but it makes the explanation more convincing.

We also investigated on the number of training examples that are required by our model to perform well. We get an accuracy of $99\%$ at worst for a couple of training images (so 33 test examples). Actually, when dealing with just one training example, since our model looks for frequent relations to set the constraints, it will extract the relations whose evaluation is larger than the thresholds we talked about in Sec.~\ref{subsec:mining}. Any example that is not an outlier should then allow the model to perform well. Thus, we show that our approach can perform spatial reasoning and achieve high accuracy from just a pair of training examples.

% \begin{figure}[tb]
%     \centering
%     \includegraphics[width=0.5\textwidth]{figures/cvpr-cropped.pdf}
%     \caption{Evolution of the accuracy with the number of training examples. We can see that the accuracy is pretty high with only one training example. When the number of training examples increases, the accuracy converges around 94.5\%.}
%     \label{fig:training}
% \end{figure}

%Based on these observations, we set the catalogue to directional, symmetry and stretched relations and and the size of the training set should be at least 20. The results we obtain are represented in Table~\ref{tab:classif}. The left kidney and the bladder are the most mislabeled organs. 

We observe that our model outperforms the CNN classifier presented in~\cite{roth2015anatomy}, which does not achieve perfect accuracy. That model was trained on a bigger training set and does not provide any kind of explanation. The closest method to ours, which was presented in~\cite{xue2017automatic}, does not give any accuracy as a baseline. Its drawback is that it can miss labels, which happens at least once every five examples. In our approach, a label cannot be missing since every variable of the FCSP has to be associated to a domain.

On a side note, the generalisability of our approach depends on how well images are segmented (although fuzzy logic helps to deal with imprecision), how expressive the vocabulary is and how many outliers are in the dataset. Applications where one of this is missing may lead to a drop in performance regarding both annotations and explanations.
%Finally, the generalisability of our model depnds on three things. First the segmentation of the input image, which is a necessary step. Images that cannot be correctly segmented may lead to a drop in performance, although fuzzy logic enables to deal with imprecision. Then, the whole approach relies on the vocabulary we use. If it is not set properly, 

% \begin{table}[tb]
% \centering
% %\resizebox{\textwidth}{!}{%
% \begin{tabular}{@{}lcc@{}}
% \toprule
% Organs       & Our model & Roth et al. \\ \midrule
% All          & 94.9      & 94.1        \\
% Liver        & 100       & 97.4        \\
% Spleen       & 97.4      & /           \\
% Right kidney & 100       & /           \\
% Left kidney  & 82.1      & /           \\
% Right lung   & 100       & 69.8        \\
% Left lung    & 100       & 69.8        \\
% Bladder      & 84.6      & /           \\ \bottomrule
% \end{tabular}%
% %}
% \caption{Comparison of our method with the baseline CNN classifier presented in~\protect\cite{roth2015anatomy}. }
% \label{tab:classif}
% \end{table}

\section{Conclusion and Prospects}
In this article, we present a novel visual learning and reasoning framework whose goal is to explain and annotate relevant objects in images. The problem is formalized as a fuzzy constraint satisfaction problem. It is based on fuzzy spatial relations, which are learned on a set of annotated objects in images and then translated into constraints. We demonstrated our approach on a medical image dataset and showed that our method takes advantage of symbolic learning and reasoning so that it explains its results and it only needs a couple of training examples to achieve $99\%$ accuracy.

In the future, we would like to work on a strategy that makes the first step of the process faster. A first idea is to determine a hierarchical structure of the spatial relations to apply a topological sort.
%A second idea is to update the order in which relations are assessed depending on previously studied examples.
%Moreover, we are working on making our model able to deal with missing organs.
Moreover, since fuzzy logic enables to manage imprecise segments, the goal is to insert an unsupervised segmentation model before the model we presented here. This would enable to adapt to different kinds of images.

Finally, this is a first step in mixing statistical machine learning (especially deep learning) for perception with symbolic learning and reasoning for higher level intelligence in order to create an explainable artificial intelligence.
%Finally, one of the main issue in the explainable artificial intelligence community is the evaluation of explanations. Finding proper ways to assess explanations is essential and is necessary for fully explainable approaches like ours.

\bibliographystyle{plain}
\bibliography{biblio}

\end{document}